\documentclass[11pt]{article}

\usepackage[margin=1in]{geometry}
\usepackage{amsmath,amssymb}
\usepackage{graphicx}
\usepackage{booktabs}
\usepackage[numbers]{natbib}
\usepackage[hidelinks]{hyperref}

\title{Time Aggregation Features for XGBoost Models}
\author{Mykola Pinchuk, PhD\thanks{Email: \texttt{pinchumkykola@gmail.com}.}\\Independent Researcher\\San Jose, USA}
\date{December 28, 2025}

\begin{document}
\maketitle

\begin{abstract}
This paper studies time aggregation features for XGBoost models in click-through rate prediction.
The setting is the Avazu click-through rate prediction dataset with strict out-of-time splits and a no-lookahead feature constraint.
Features for hour $H$ use only impressions from hours strictly before $H$.
This paper compares a strong time-aware target encoding baseline to models augmented with entity history time aggregation under several window designs.
Across two rolling-tail folds on a deterministic ten percent sample, a trailing window specification improves ROC AUC by about 0.0066 to 0.0082 and PR AUC by about 0.0084 to 0.0094 relative to target encoding alone.
Within the time aggregation design grid, event count windows provide the only consistent improvement over trailing windows, and the gain is small.
Gap windows and bucketized windows underperform simple trailing windows in this dataset and protocol.
These results support a practical default of trailing windows, with an optional event count window when marginal ROC AUC gains matter. Code to reproduce all results is available at \url{https://github.com/MykolaPinchuk/paper_time_aggregations}.

\end{abstract}

\section{Introduction}
Many industrial prediction tasks are temporally structured.
The data distribution drifts, and high cardinality identifiers can repeat.
These properties make entity history features attractive.
They also make evaluation fragile.
If features incorporate future information, even in subtle ways, reported gains can be illusory.

This paper focuses on time aggregation features for XGBoost models.
It studies simple entity history statistics computed over recent windows.
The paper uses strict out-of-time splits and a no-lookahead feature constraint in which features for hour $H$ use only events from hours $< H$.

The main contribution is empirical.
This paper shows that no-lookahead time aggregation yields a clear lift over a time-aware target encoding baseline.
It then quantifies how performance changes under window design choices.
In this setting, trailing windows are a strong default.
Event count windows are the most reliable modification.
Gap windows and bucketized windows are consistently worse.

Section 2 defines the dataset and evaluation protocol.
Section 3 defines the feature families and window constructions.
Section 4 reports results and practical guidance.

\section{Related Work}
High cardinality tabular prediction tasks arise across domains.
They often combine sparsity, repeated identifiers, and temporal drift.
The Avazu click-through rate benchmark is one public example.
In this setting, industrial systems have historically relied on linear models trained with online optimization and feature hashing \citep{mcmahan2013ad}.
More expressive approaches include factorization machines and deep models that combine memorization and generalization \citep{rendle2010factorization,cheng2016wide,guo2017deepfm}.

Gradient boosted decision trees are a strong baseline for tabular data and remain widely used in industry \citep{friedman2001greedy,chen2016xgboost}.
Their performance depends heavily on feature engineering.
In particular, history-based count and rate features are common because identifiers repeat and the data distribution drifts.
Industry studies also emphasize the importance of time-aware evaluation and deployment constraints when comparing feature sets \citep{he2014practical}.

History-based features can leak future information if aggregates are computed using data from the prediction horizon.
Out-of-time splitting and no-lookahead feature simulation are common safeguards \citep{he2014practical}.
Target encoding is another strong technique for high cardinality categories.
It requires careful fold based estimation to avoid target leakage \citep{micci2001preprocessing}.
This paper focuses on simple, no-lookahead time aggregation features and quantifies how design choices affect performance under out-of-time evaluation.

\section{Problem Setup}
This paper uses the Kaggle Avazu click-through rate prediction dataset.
Each row is an ad impression, meaning a single event in which an advertisement is shown to a user.
The label indicates whether the shown ad was clicked.
Each row also includes a timestamp at hour resolution.
All time windows in this paper are defined in hours.
The feature space contains many high cardinality categorical fields such as device and site identifiers.

The prediction task is to estimate the probability of click for each impression.
This paper reports ROC AUC as the primary metric and PR AUC as a secondary metric.

Evaluation uses strict out-of-time splits.
This study uses a deterministic ten percent sample and two rolling-tail folds.
Each fold uses three disjoint, chronologically ordered splits.
Fold A trains on days up to $D-2$, validates on day $D-1$, and tests on day $D$.
Fold B trains on days up to $D-3$, validates on day $D-2$, and tests on day $D-1$.
The ten percent sample is constructed deterministically from the row identifier using a hash-based filter.
Appendix Section \ref{sec:appendix_sampling} provides the exact rule.
In this sample, the date range spans 2014-10-21 through 2014-10-30 and contains about 4.04 million impressions.
Table \ref{tab:splits_10pct} reports the exact split ranges, row counts, and click rates.

\begin{table}[t]
  \centering
  \small
  \begin{tabular}{lllllll}
\toprule
sample & fold\_id & split & start\_hour & end\_hour & n\_rows & click\_rate \\
\midrule
10pct & A & test & 2014-10-30 00:00 & 2014-10-30 23:00 & 421,303 & 0.1691 \\
10pct & A & train & 2014-10-21 00:00 & 2014-10-28 23:00 & 3,238,511 & 0.1715 \\
10pct & A & val & 2014-10-29 00:00 & 2014-10-29 23:00 & 382,897 & 0.1570 \\
10pct & B & test & 2014-10-29 00:00 & 2014-10-29 23:00 & 382,897 & 0.1570 \\
10pct & B & train & 2014-10-21 00:00 & 2014-10-27 23:00 & 2,708,333 & 0.1752 \\
10pct & B & val & 2014-10-28 00:00 & 2014-10-28 23:00 & 530,178 & 0.1525 \\
\bottomrule
\end{tabular}

  \caption{Exact rolling-tail splits on the deterministic ten percent sample, including hour ranges, number of rows, and click rates.}
  \label{tab:splits_10pct}
\end{table}

\begin{figure}[t]
  \centering
  \includegraphics[width=\linewidth]{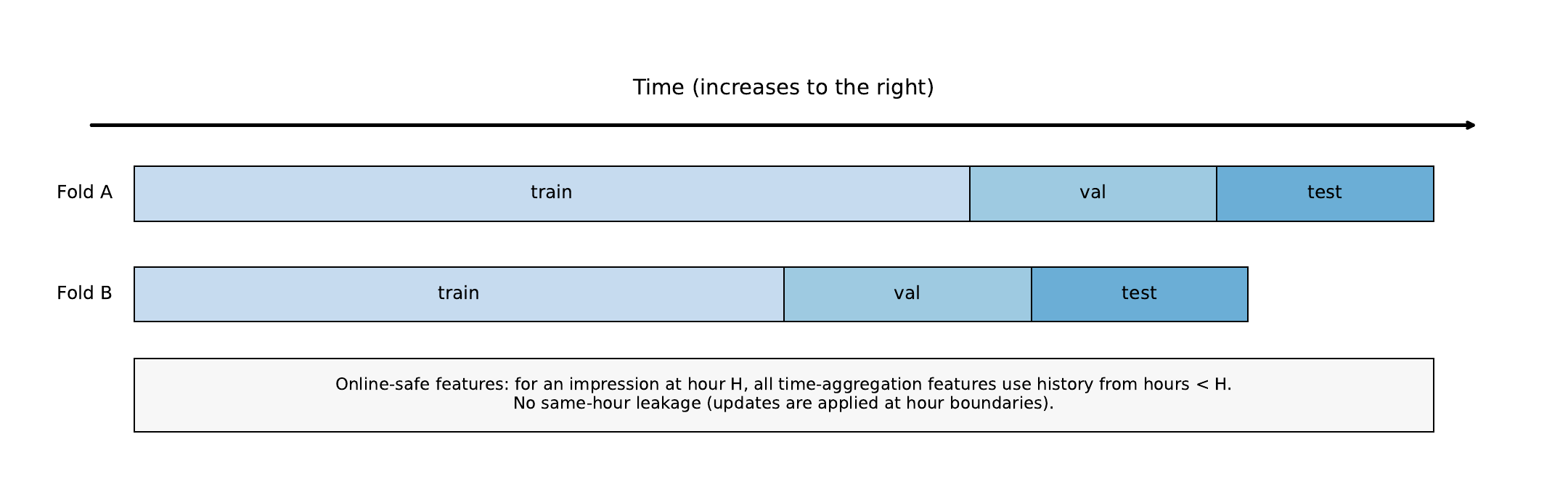}
  \caption{Evaluation protocol and no-lookahead feature constraint. For an impression at hour $H$, history-based features use only events from hours $< H$, and they exclude same-hour information.}
  \label{fig:protocol}
\end{figure}

Feature computation follows a no-lookahead constraint.
For impressions at hour $H$, aggregation features and encodings may use only events from hours strictly before $H$.
This excludes same-hour information to avoid leakage through contemporaneous counts.
Figure \ref{fig:protocol} summarizes the evaluation protocol and the feature availability constraint.

\section{Method}
The model is an XGBoost classifier \citep{chen2016xgboost}.
Hyperparameters are fixed across experiments.
Training uses early stopping on the validation split.

Appendix Table \ref{tab:xgb_params} lists key model hyperparameters.
The model uses objective \texttt{binary:logistic}.
Training uses early stopping rounds $=50$ with evaluation metrics AUC and AUC PR on the validation split.

This paper compares three feature configurations.
The first configuration is a time-aware target encoding baseline.
The second configuration adds time aggregation features to this baseline.
The third configuration removes target encoding and uses only time aggregation features as a sensitivity check.
Target encoding follows standard practice for high cardinality categorical features \citep{micci2001preprocessing}.

The time-aware target encoding baseline is computed at hour resolution with a one-hour shift.
For a row at hour $h$, statistics use only labels from hours $< h$.
Appendix Section \ref{sec:appendix_te_details} provides the full definition and implementation details.

Time aggregation features are entity history summaries computed under the no-lookahead constraint.
Time is represented at hour resolution.
For an impression at hour $H$, all history-based features use only events from hours strictly before $H$.
This excludes same-hour information.

The following entity keys are used: device ip, device id, app id, and site id.
For each key and each window, the features include impression counts and smoothed click rates.
For impressions $I$ and clicks $C$ in a window, the impression feature is $\log(1+I)$.
The click rate feature is $(C + \alpha) / (I + \alpha + \beta)$ with $\alpha=1$ and $\beta=10$.

This paper varies two aspects of window design.
The first aspect is the length tuple, expressed in hours back.
The second aspect is the window shape.
The window shapes include trailing windows, a one-hour gap variant, bucketized recency buckets, calendar-aligned windows, and an event count window defined as the last fifty impressions.

This paper defines a window specification as a length tuple and a shape.
For example, the trailing specification with length tuple $(1, 6, 24, 48, 168)$ uses trailing windows at those hour lengths.

\subsection{Window length tuples}
Table \ref{tab:length_sets} lists the benchmarked length tuples.
Each tuple is evaluated under multiple window shapes.

\begin{table}[t]
  \centering
  \begin{tabular}{l}
    \toprule
    Length tuple (hours) \\
    \midrule
    $(1, 6, 24)$ \\
    $(1, 3, 6, 12, 24)$ \\
    $(1, 6, 24, 48, 168)$ \\
    $(1, 2, 4, 8, 16, 24, 48, 96, 168)$ \\
    \bottomrule
  \end{tabular}
  \caption{Window length tuples used in the design grid.}
  \label{tab:length_sets}
\end{table}

\subsection{Window shapes}
Window shapes define how the history for an impression at hour $H$ is converted into time windows.
For a window of length $L$ hours, a trailing window uses the half open interval $[H-L, H)$.

Table \ref{tab:window_shapes} summarizes the benchmarked shapes.
Shapes that ``add'' windows are applied on top of trailing windows for the same length tuple.

\begin{table}[t]
  \centering
  \begin{tabular}{p{0.18\linewidth}p{0.76\linewidth}}
    \toprule
    Shape & Definition used in this study \\
    \midrule
    trailing &
      For each $L$ in the length tuple, use events in $[H-L, H)$. \\
    gap1 &
      For each $L$ in the length tuple, use events in $[H-(L+1), H-1)$.
      This excludes the most recent hour. \\
    bucket &
      Let $0 < e_1 < \dots < e_K$ be the hour edges from the length tuple.
      Construct disjoint buckets $[H-e_1, H)$, $[H-e_2, H-e_1)$, \dots, $[H-e_K, H-e_{K-1})$.
      Features are computed per bucket. \\
    calendar &
      Add calendar aligned day windows on top of trailing windows.
      For day based lengths, such as $L \in \{24, 48, 168\}$, the window is aligned to full days strictly before the impression hour.
      For sub day lengths, the calendar window matches the trailing window. \\
    event50 &
      Add an event count window on top of trailing windows.
      For each entity, use the most recent 50 impressions strictly before hour $H$. \\
    \bottomrule
  \end{tabular}
  \caption{Window shape definitions.}
  \label{tab:window_shapes}
\end{table}

\section{Experiments}
This section reports empirical results under the evaluation protocol described above.
All experiments share the same data splits, model family, entity set, and training protocol.

Time aggregation yields a clear lift over a target-encoding-only baseline under the no-lookahead protocol.
Appendix Figure \ref{fig:te_lift_deltas} reports paired deltas with confidence intervals for this comparison.
Table \ref{tab:absolute_metrics_selected} reports the corresponding absolute test metrics for selected specifications.
Appendix Table \ref{tab:top_specs_absolute} reports additional absolute metrics for representative high-performing specifications.

\begin{table}[t]
  \centering
  \small
  \resizebox{\linewidth}{!}{%
  \begin{tabular}{lcccccc}
    \toprule
    & \multicolumn{3}{c}{Test ROC AUC} & \multicolumn{3}{c}{Test PR AUC} \\
    \cmidrule(lr){2-4} \cmidrule(lr){5-7}
    Specification & A & B & mean & A & B & mean \\
    \midrule
    TE only &
      0.74175 & 0.74382 & 0.74278 &
      0.36931 & 0.34796 & 0.35863 \\
    TE plus trailing $(1, 6, 24, 48, 168)$ &
      0.74831 & 0.75198 & 0.75014 &
      0.37875 & 0.35629 & 0.36752 \\
    TE plus trailing $(1, 6, 24, 48, 168)$ plus event50 &
      0.74864 & 0.75245 & 0.75054 &
      0.37844 & 0.35745 & 0.36794 \\
    \bottomrule
  \end{tabular}
  }
  \caption{Absolute test ROC AUC and PR AUC on Fold A and Fold B for selected specifications under the TE plus time aggregation setting.}
  \label{tab:absolute_metrics_selected}
\end{table}

\subsection{Window design effects without target encoding}
\begin{figure}[t]
  \centering
  \includegraphics[width=\linewidth]{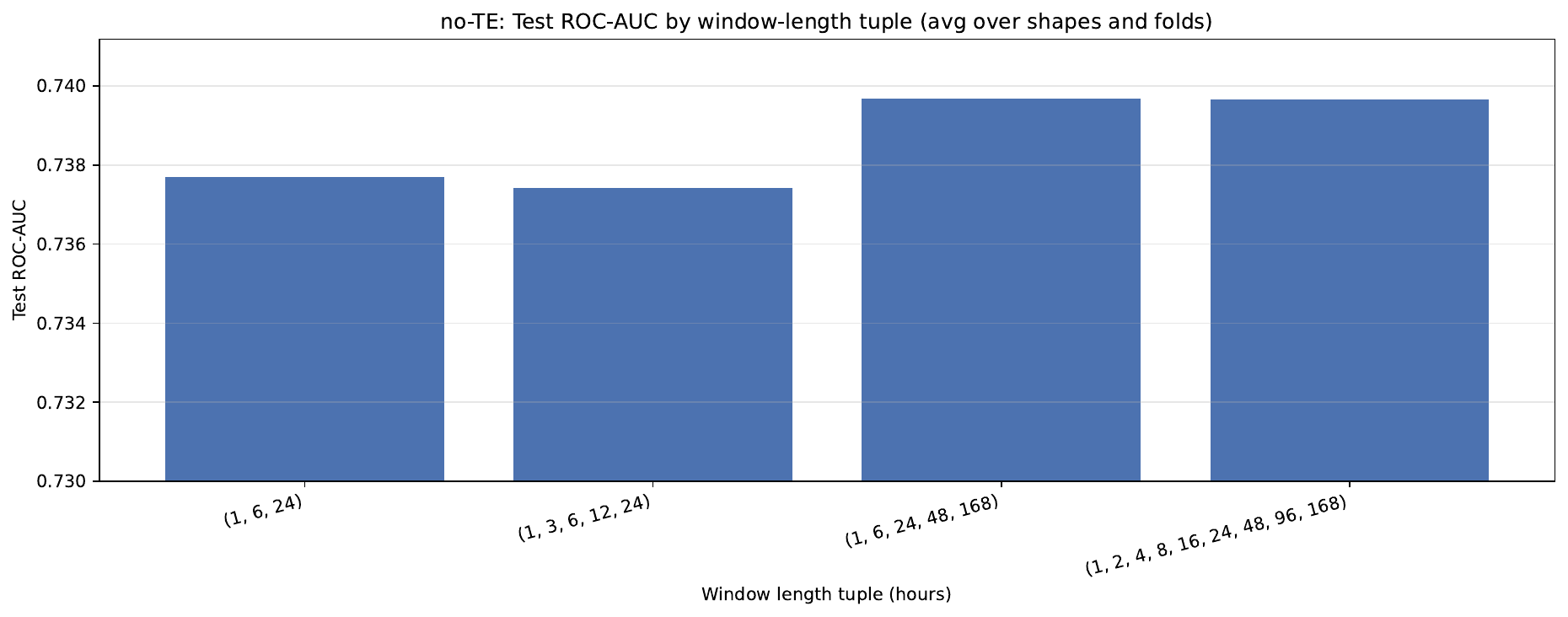}
  \vspace{0.5em}
  \includegraphics[width=\linewidth]{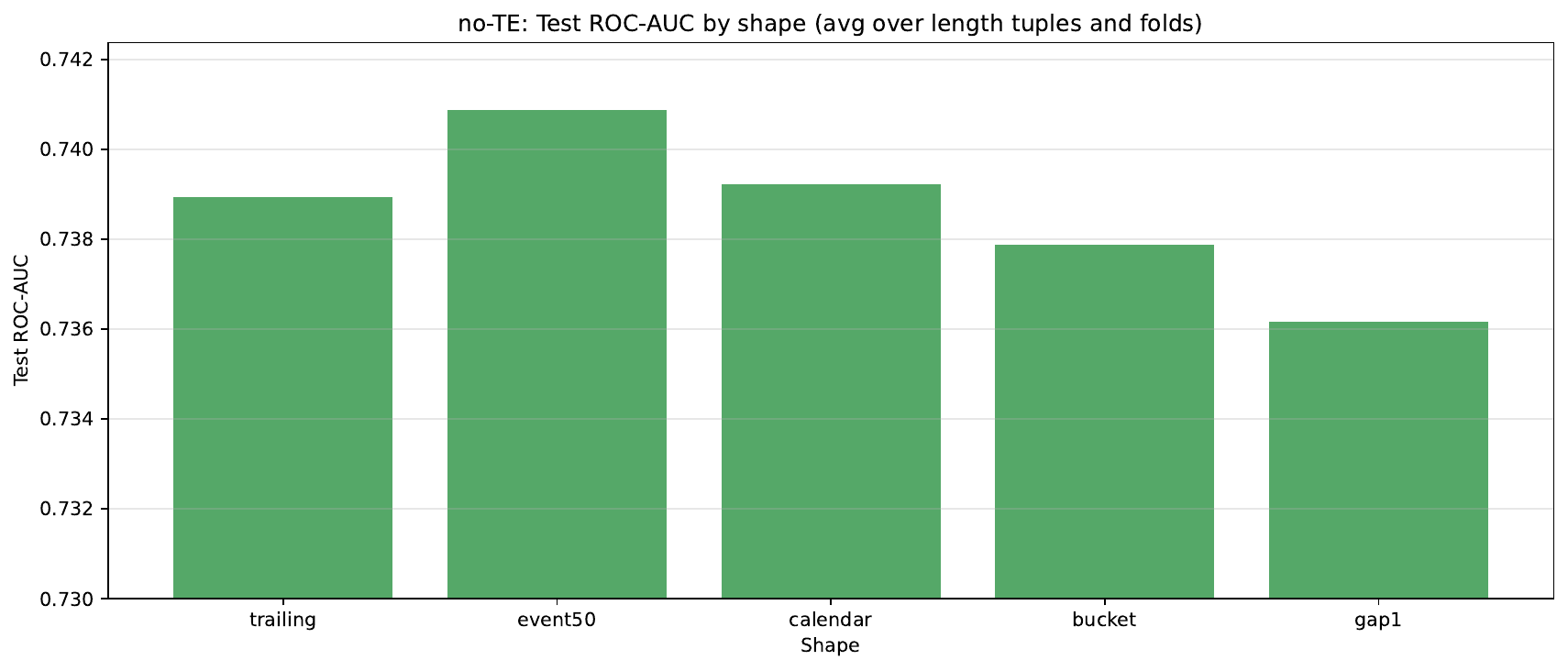}
  \caption{No target encoding sensitivity view. Panel A reports test ROC AUC by window length tuple, averaged over shapes and folds. Panel B reports test ROC AUC by window shape, averaged over length tuples and folds.}
  \label{fig:trimmed_note_rocauc}
\end{figure}

Figure \ref{fig:trimmed_note_rocauc} provides a compact view of window design effects in the setting without target encoding.
It reports test ROC AUC averaged over folds, and it marginalizes over the other design dimension.

\subsection{Window design summary}
\begin{figure}[t]
  \centering
  \includegraphics[width=\linewidth]{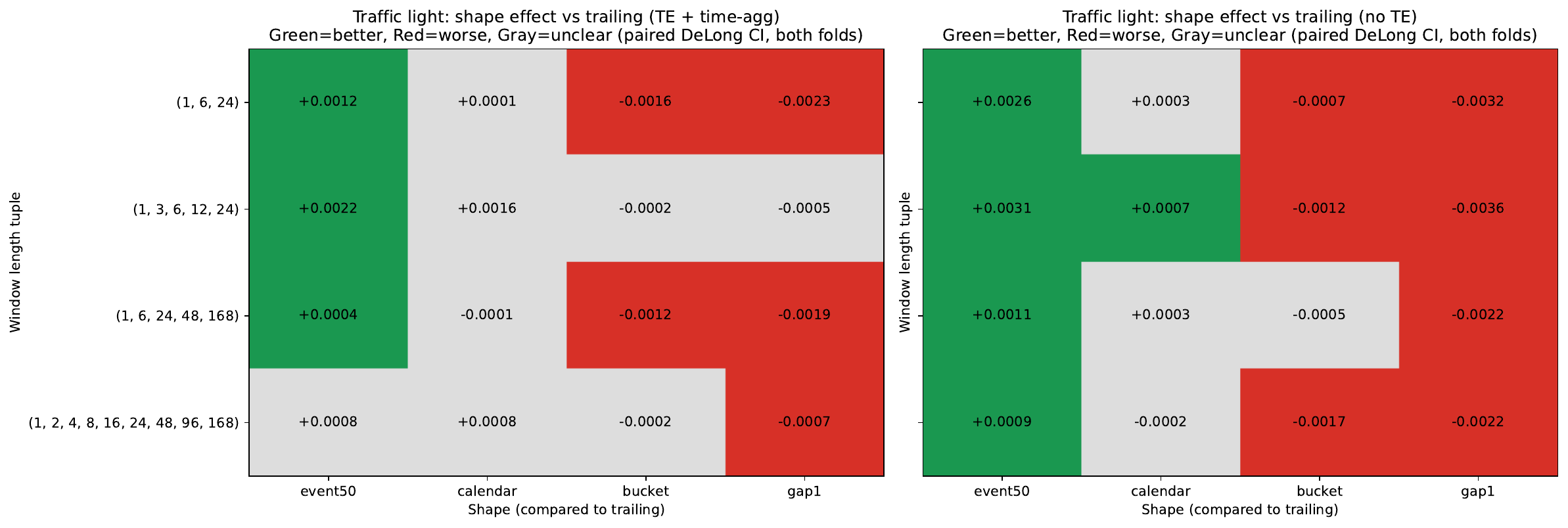}
  \caption{Traffic light summary of shape effects relative to trailing windows, split by whether target encoding is used.}
  \label{fig:traffic_light}
\end{figure}

Figure \ref{fig:traffic_light} summarizes shape effects relative to trailing windows within each length tuple.
Event count windows are positive in most settings.
Gap and bucket shapes are negative in most settings.
Calendar alignment effects are small and mixed.

\subsection{Window design choices with target encoding}
The next comparison evaluates window design choices with target encoding enabled under the same evaluation protocol.
Trailing windows with length tuple $(1, 6, 24, 48, 168)$ are a strong baseline.
Event count windows are the only consistently helpful shape modification.
Gap windows and bucketized windows underperform trailing windows within the same length tuple.

\begin{figure}[t]
  \centering
  \includegraphics[width=\linewidth]{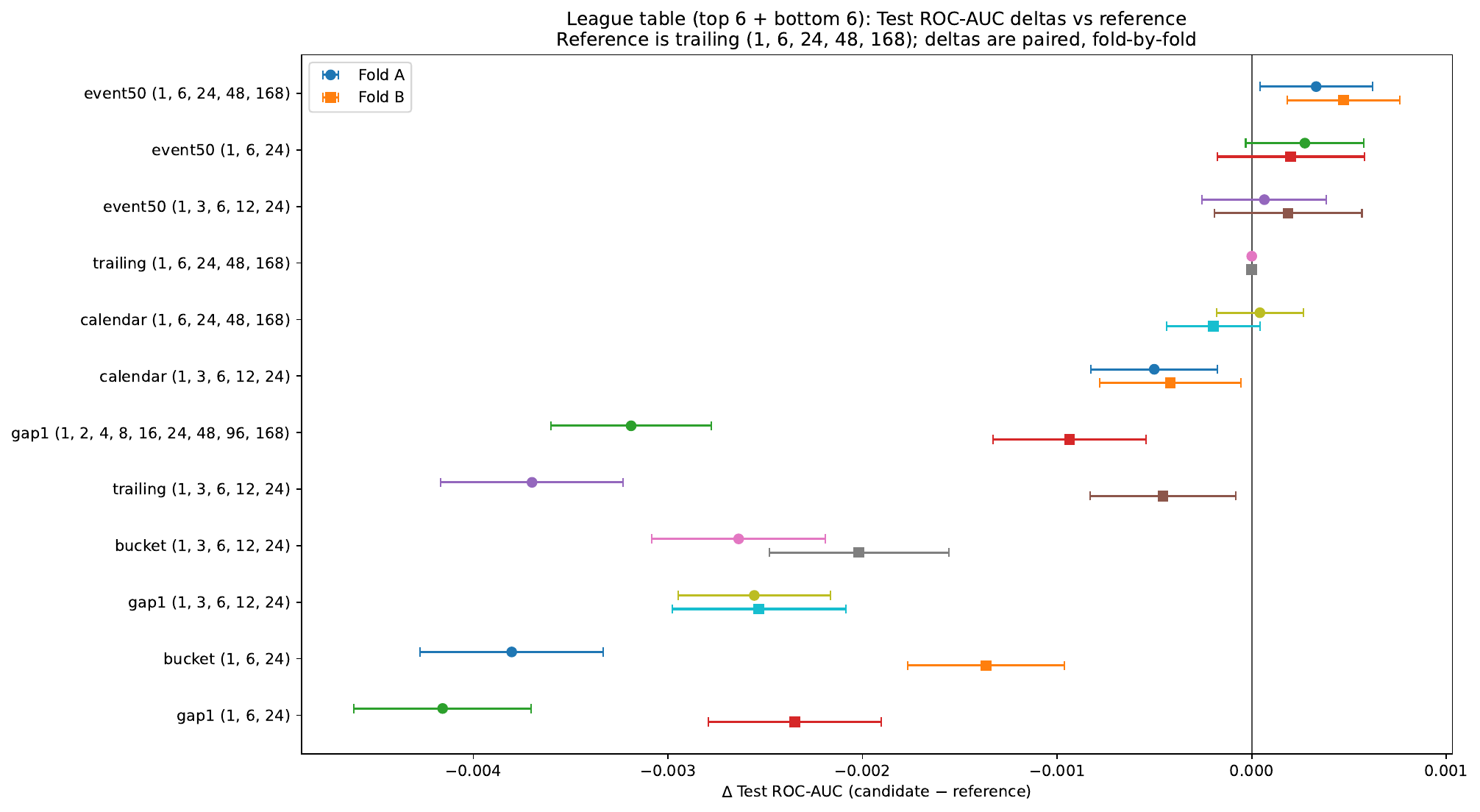}
  \caption{League table of best and worst specifications by paired deltas in test ROC AUC relative to trailing $(1, 6, 24, 48, 168)$.}
  \label{fig:league_table}
\end{figure}

Figure \ref{fig:league_table} provides a ranked view of best and worst specifications using paired deltas in test ROC AUC.

Appendix Figure \ref{fig:decision_plot} provides a detailed view that separates the effect of length tuple choices from the effect of shape choices.
It reports paired deltas and fold specific estimates.

\subsection{Interaction with target encoding}
Target encoding yields a large uplift relative to time aggregation only models in this setup.
Appendix Figure \ref{fig:te_uplift} reports the uplift across specifications.
Even with target encoding, event count windows provide a small additional gain in ROC AUC over trailing windows.
The gain is $0.00033$ on Fold A and $0.00047$ on Fold B in the paired tests.
PR AUC gains from event count windows are small and mixed.

\subsection{Practical guidance}
For this dataset and protocol, a trailing window design is a robust default.
An event count window can be added when a small additional ROC AUC gain is valuable.
Gap windows and bucketized windows are not recommended under no-lookahead evaluation.

\clearpage

\section{Discussion}
This section discusses potential mechanisms behind the observed window length and shape effects.
The discussion is hypothesis driven.
The goal is to guide future experiments that test these mechanisms directly.

Length tuples control which time scales are exposed to the model.
Click behavior can change over minutes, hours, and days.
Short windows can capture immediate recency effects and repeated exposure.
Day and week scale windows can capture periodicity and slower drift.
If a length tuple omits a relevant scale, a different window shape cannot recover it.
In that case, shape changes mainly redistribute the same historical mass.

Trailing windows represent recent history and its decay in a direct way.
They also provide a simple bias variance tradeoff.
As $L$ increases, a trailing window reduces variance of rate estimates but can increase bias under drift.
A small set of trailing windows creates a multi resolution summary.
A tree model can combine these summaries in a nonlinear way.
This can be sufficient when the main signal is recency and repetition.

Gap windows exclude the most recent hour from the history window.
This can remove the most informative part of the signal when behavior changes quickly.
It can also remove short term dependence for repeated identifiers.
Gap windows can be useful when very recent data is contaminated by a leakage channel.
In this study, the no-lookahead constraint already excludes same-hour leakage.
Therefore, the gap mainly acts as an unnecessary information restriction.

Bucketized windows partition history into disjoint intervals.
This increases feature dimensionality and can increase estimation variance.
Each bucket contains fewer events than an overlapping trailing window at a similar horizon.
Bucketization also removes cumulative smoothing from nested trailing windows.
If the signal is approximately monotone in recency, trailing windows represent it efficiently.
Bucketization may help when distinct time bands have distinct effects.

Fixed length time windows implicitly assume a stable event rate.
For many entity keys, event rates vary widely across entities and over time.
An event count window adapts by fixing the number of observations used to estimate a rate.
This can stabilize estimates for sparse entities.
It can also reduce the impact of bursty traffic for frequent entities.
It may also be more robust under hour resolution timestamps with irregular sampling.

Calendar aligned windows aim to represent full previous days or weeks.
They can help when boundary effects exist at calendar transitions.
However, in hour resolution data, calendar alignment can be redundant with explicit time features.
Examples include hour of day and day of week.
Calendar blocks can also mix heterogeneous periods under drift.
These factors can yield small effects with signs that vary across length tuples.

\section{Conclusion}
This paper evaluates time aggregation features for XGBoost models under a no-lookahead out-of-time protocol on Avazu.
Entity history aggregation yields a clear lift over a target-encoding-only baseline.
Within a controlled design grid, trailing windows are a strong default.
Event count windows provide a small and consistent improvement in ROC AUC.
Gap and bucket window constructions are consistently worse in this setting.

These results are limited to one dataset and a small set of folds.
The evaluation is leakage averse, but it is not a full time series cross validation study.
Future work should validate the conclusions on additional datasets and under different drift regimes.

\bibliographystyle{plainnat}
\bibliography{references}

@inproceedings{chen2016xgboost,
  title     = {XGBoost: A Scalable Tree Boosting System},
  author    = {Chen, Tianqi and Guestrin, Carlos},
  booktitle = {Proceedings of the 22nd ACM SIGKDD International Conference on Knowledge Discovery and Data Mining},
  year      = {2016}
}

@inproceedings{micci2001preprocessing,
  title     = {A Preprocessing Scheme for High Cardinality Categorical Attributes in Classification and Prediction Problems},
  author    = {Micci-Barreca, Daniele},
  booktitle = {ACM SIGKDD Explorations Newsletter},
  year      = {2001}
}

@article{friedman2001greedy,
  title   = {Greedy Function Approximation: A Gradient Boosting Machine},
  author  = {Friedman, Jerome H.},
  journal = {The Annals of Statistics},
  year    = {2001},
  volume  = {29},
  number  = {5},
  pages   = {1189--1232}
}

@inproceedings{mcmahan2013ad,
  title     = {Ad Click Prediction: a View from the Trenches},
  author    = {McMahan, H. Brendan and Holt, Gary and Sculley, D. and Young, Michael and Ebner, Dietmar and Grady, Julian and Nie, Lan and Phillips, Todd and Davydov, Eugene and Golovin, Daniel and Chikkerur, Sharat and Liu, Dan and Wattenberg, Martin and Hrafnkelsson, Arnar and Boulos, Tom and Kubica, Jeremy},
  booktitle = {Proceedings of the 19th ACM SIGKDD International Conference on Knowledge Discovery and Data Mining},
  year      = {2013},
  pages     = {1222--1230}
}

@inproceedings{he2014practical,
  title     = {Practical Lessons from Predicting Clicks on Ads at Facebook},
  author    = {He, Xinran and Pan, Junfeng and Jin, Ou and Xu, Tianbing and Liu, Bo and Xu, Tao and Shi, Yanxin and Atallah, Antoine and Herbrich, Ralf and Bowers, Stuart and Quinonero-Candela, Joaquin},
  booktitle = {Proceedings of the Eighth International Workshop on Data Mining for Online Advertising},
  year      = {2014}
}

@inproceedings{rendle2010factorization,
  title     = {Factorization Machines},
  author    = {Rendle, Steffen},
  booktitle = {2010 IEEE International Conference on Data Mining},
  year      = {2010},
  pages     = {995--1000}
}

@article{cheng2016wide,
  title   = {Wide \& Deep Learning for Recommender Systems},
  author  = {Cheng, Heng-Tze and Koc, Levent and Harmsen, Jeremiah and Shaked, Tal and Chandra, Tushar and Aradhye, Hrishi and Anderson, Glen and Corrado, Greg and Chai, Wei and Ispir, Mustafa and Anil, Rohan and Haque, Zakaria and Hong, Lichan and Jain, Vihan and Liu, Xiaobing and Shah, Hemal},
  journal = {arXiv preprint arXiv:1606.07792},
  year    = {2016}
}

@inproceedings{guo2017deepfm,
  title     = {DeepFM: A Factorization-Machine based Neural Network for CTR Prediction},
  author    = {Guo, Huifeng and Tang, Ruiming and Ye, Yunming and Li, Zhenguo and He, Xiuqiang},
  booktitle = {Proceedings of the 26th International Joint Conference on Artificial Intelligence},
  year      = {2017}
}

\clearpage
\appendix
\section{Additional Details}
\subsection*{Dataset and application context}
Click-through rate prediction is one example of a large-scale, temporally structured, tabular classification problem.
An advertisement is shown to a user in response to a page view or an app interaction.
Each showing is an impression.
Some impressions lead to a click.
The click-through rate is the probability of click conditional on an impression.
Accurate click probability estimates are used in ranking and bidding pipelines.

The Avazu dataset is a public snapshot of an ad marketplace log at impression level.
Each row corresponds to one impression and includes a binary click label and a timestamp at hour resolution.
The feature space is dominated by high cardinality categorical identifiers that describe the user device, the application, and the site.
These identifiers induce strong repetition patterns and create a large cold-start surface as new values appear over time.
The label is sparse, so PR AUC complements ROC AUC by emphasizing precision and recall under class imbalance.

This setting is temporally nonstationary.
Traffic composition and user intent change over time, and the statistical relationship between identifiers and click probability drifts.
Therefore, offline evaluation must be time-aware.
This paper uses strict out-of-time splits and a no-lookahead simulation of feature availability.
These constraints aim to prevent leakage and to better approximate a production deployment where features must be computed from past events only.

\subsection{Deterministic sampling procedure}
\label{sec:appendix_sampling}
The ten percent sample is constructed deterministically by applying a hash filter to the Avazu row identifier while streaming the dataset.
For each row, the identifier is hashed with the pandas function \texttt{hash\_pandas\_object} and interpreted as an unsigned 64-bit integer $h$.
The row is included in the ten percent sample if $(h \bmod 100) < 10$.
This procedure is deterministic within a fixed pandas version.
Pandas does not guarantee cross-version hash stability, so exact sample identity may change across pandas versions.

\subsection*{Model hyperparameters}
Table \ref{tab:xgb_params} reports XGBoost hyperparameters used in all experiments.

\begin{table}[t]
  \centering
  \begin{tabular}{ll}
    \toprule
    Hyperparameter & Value \\
    \midrule
    $n\_estimators$ & 800 \\
    $learning\_rate$ & 0.05 \\
    $max\_depth$ & 6 \\
    $subsample$ & 0.8 \\
    $colsample\_bytree$ & 0.8 \\
    $min\_child\_weight$ & 10.0 \\
    $reg\_lambda$ & 5.0 \\
    $random\_state$ & 42 \\
    $n\_jobs$ & 1 \\
    $tree\_method$ & \texttt{hist} \\
    \bottomrule
  \end{tabular}
  \caption{XGBoost hyperparameters used in all experiments.}
  \label{tab:xgb_params}
\end{table}

\subsection{Time-aware target encoding}
\label{sec:appendix_te_details}
This section specifies the target encoding baseline used in the paper.
Target encoding is computed at hour resolution with a one-hour shift.
For a row at hour $h$, statistics use only labels from hours $< h$.

\subsubsection*{Split-boundary handling}
For each fold, target encoding and time aggregation features are computed for all rows in chronological order before slicing into train, validation, and test splits.
Therefore, features in the validation and test splits may depend on label feedback from earlier hours in the same fold, including earlier hours within the same day, but never from the current hour.
This matches a streaming evaluation setting in which click outcomes become available with a one-hour delay.

The model is trained on numeric features only.
It includes:
\begin{itemize}
  \item base time features: hour of day and the global prior $\mathrm{prior\_ctr}(h)$,
  \item for each categorical column $c$: a target encoding feature $c\_\_te$ and a history volume feature $c\_\_hist\_imps=\log(1+I_{v,<h})$.
\end{itemize}

Table \ref{tab:cat_cols} lists the categorical columns that are target encoded in the ten percent sample.

\begin{table}[t]
  \centering
  \small
  \begin{tabular}{lll}
    \toprule
    \multicolumn{3}{c}{Categorical columns} \\
    \midrule
    C1 & banner\_pos & site\_id \\
    site\_domain & site\_category & app\_id \\
    app\_domain & app\_category & device\_id \\
    device\_ip & device\_model & device\_type \\
    device\_conn\_type & C14 & C15 \\
    C16 & C17 & C18 \\
    C19 & C20 & C21 \\
    \bottomrule
  \end{tabular}
  \caption{Categorical columns target encoded in the ten percent sample.}
  \label{tab:cat_cols}
\end{table}

The global time-aware prior CTR is
\[
\mathrm{prior\_ctr}(h) = \frac{C_{<h} + a}{I_{<h} + a + b},
\]
with $a=1$ and $b=10$, where $I_{<h}$ and $C_{<h}$ are cumulative impressions and clicks over hours $< h$.

For a categorical column $c$ and value $v$, define cumulative counts $I_{v,<h}$ and $C_{v,<h}$ over hours $<h$.
The target encoding value is
\[
\mathrm{TE}_c(v,h) = \frac{C_{v,<h} + m \cdot \mathrm{prior\_ctr}(h)}{I_{v,<h} + m},
\]
with smoothing strength $m=100$.
If a value is unseen, $I_{v,<h}=0$ and the encoder returns $\mathrm{prior\_ctr}(h)$.
In this case, $c\_\_hist\_imps=\log(1+0)=0$.

The algorithm is deterministic and can be summarized as follows.
\begin{enumerate}
  \item Sort rows by hour in ascending order.
  \item For each hour $h$, compute $\mathrm{prior\_ctr}(h)$ from cumulative counts over hours $<h$.
  \item For each categorical column $c$ and each row at hour $h$ with value $v$, compute $c\_\_te$ and $c\_\_hist\_imps$ from $I_{v,<h}$ and $C_{v,<h}$.
  \item After processing hour $h$, update cumulative counts with all rows from hour $h$.
\end{enumerate}

\begin{figure}[t]
  \centering
  \includegraphics[width=\linewidth]{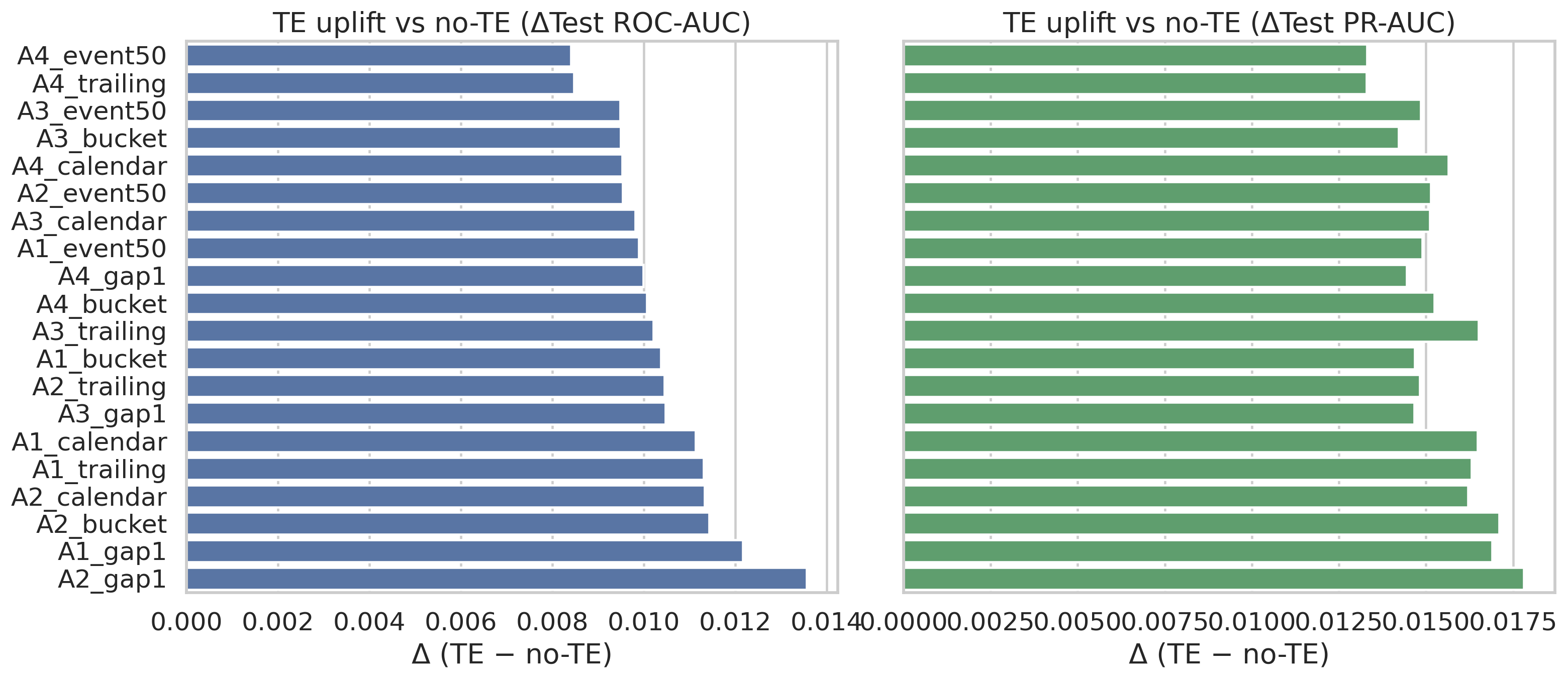}
  \caption{Uplift from adding target encoding on top of time aggregation features.}
  \label{fig:te_uplift}
\end{figure}

\subsection{Time aggregation feature definitions}
\label{sec:appendix_timeagg_defs}
Time aggregation features are computed online-safe with per-hour batching.
For a row at hour $h$, all entity-history statistics use only data from hours $<h$.
Each entity feature uses impression and click counts in a window, reported as $\log(1+I)$ and a smoothed rate $(C+\alpha)/(I+\alpha+\beta)$ with $\alpha=1$ and $\beta=10$.

\paragraph{Trailing windows.}
For entity column $e$ and entity value $v$, and window length $w$ hours, define impressions $I_{v,[h-w,h)}$ and clicks $C_{v,[h-w,h)}$ over $[h-w,h)$.
The features are $\log(1+I_{v,[h-w,h)})$ and $(C_{v,[h-w,h)}+\alpha)/(I_{v,[h-w,h)}+\alpha+\beta)$.

\paragraph{Gap windows.}
For gap size $g$, define $[h-w-g,h-g)$ and compute counts as differences of trailing windows.

\paragraph{Bucket windows.}
Given edges $0<b_1<\dots<b_K$, bucket $j$ is $[h-b_j,h-b_{j-1})$ with $b_0=0$.
Counts are computed as differences of trailing windows at horizons $b_j$ and $b_{j-1}$.

\paragraph{Calendar windows.}
Calendar features include counts and smoothed rates for the current day up to hour $h$ and, when available, for the previous day.

\paragraph{Recency.}
For each entity value, recency is the time since last observation in hours.
Recency is undefined for unseen entities and is represented as NaN.

\paragraph{Event-count window.}
For each entity value, the event-count window uses the most recent $N$ impressions strictly before hour $h$.
When an entity has fewer than $N$ prior impressions, the features are computed from its available history, and the $\alpha,\beta$ smoothing prevents extreme rates.
In this paper, the event-count shape uses $N=50$.

\subsection{Feature dimensionality}
\label{sec:appendix_feature_dimensionality}
Table \ref{tab:feature_counts} reports total feature counts for each length tuple and shape in the target encoding grid.

\begin{table}[t]
  \centering
  \small
  \begin{tabular}{llr}
\toprule
Length tuple (hours) & Shape & Total features \\
\midrule
$(1, 6, 24)$ & trailing & 72 \\
$(1, 6, 24)$ & gap1 & 72 \\
$(1, 6, 24)$ & bucket & 72 \\
$(1, 6, 24)$ & calendar & 88 \\
$(1, 6, 24)$ & event50 & 80 \\
$(1, 3, 6, 12, 24)$ & trailing & 88 \\
$(1, 3, 6, 12, 24)$ & gap1 & 88 \\
$(1, 3, 6, 12, 24)$ & bucket & 88 \\
$(1, 3, 6, 12, 24)$ & calendar & 104 \\
$(1, 3, 6, 12, 24)$ & event50 & 96 \\
$(1, 6, 24, 48, 168)$ & trailing & 88 \\
$(1, 6, 24, 48, 168)$ & gap1 & 88 \\
$(1, 6, 24, 48, 168)$ & bucket & 88 \\
$(1, 6, 24, 48, 168)$ & calendar & 104 \\
$(1, 6, 24, 48, 168)$ & event50 & 96 \\
$(1, 2, 4, 8, 16, 24, 48, 96, 168)$ & trailing & 120 \\
$(1, 2, 4, 8, 16, 24, 48, 96, 168)$ & gap1 & 120 \\
$(1, 2, 4, 8, 16, 24, 48, 96, 168)$ & bucket & 120 \\
$(1, 2, 4, 8, 16, 24, 48, 96, 168)$ & calendar & 136 \\
$(1, 2, 4, 8, 16, 24, 48, 96, 168)$ & event50 & 128 \\
\bottomrule
\end{tabular}

  \caption{Total number of model input features for each window specification in the target encoding plus time aggregation grid.}
  \label{tab:feature_counts}
\end{table}

\subsection{Additional absolute metrics}
\label{sec:appendix_top_specs_absolute_metrics}
Table \ref{tab:top_specs_absolute} reports absolute test metrics for representative high-performing specifications in the target encoding plus time aggregation grid.

\begin{table}[t]
  \centering
  \small
  \resizebox{\linewidth}{!}{%
  \begin{tabular}{llcccc}
\toprule
Length tuple (hours) & Shape & Test ROC AUC A & Test ROC AUC B & Test PR AUC A & Test PR AUC B \\
\midrule
$(1, 6, 24)$ & event50 & 0.7486 & 0.7522 & 0.3774 & 0.3576 \\
$(1, 3, 6, 12, 24)$ & calendar & 0.7478 & 0.7516 & 0.3766 & 0.3566 \\
$(1, 3, 6, 12, 24)$ & event50 & 0.7484 & 0.7522 & 0.3788 & 0.3571 \\
$(1, 6, 24, 48, 168)$ & calendar & 0.7483 & 0.7518 & 0.3774 & 0.3564 \\
$(1, 6, 24, 48, 168)$ & event50 & 0.7486 & 0.7524 & 0.3784 & 0.3574 \\
$(1, 6, 24, 48, 168)$ & trailing & 0.7483 & 0.7520 & 0.3787 & 0.3563 \\
$(1, 2, 4, 8, 16, 24, 48, 96, 168)$ & calendar & 0.7473 & 0.7519 & 0.3772 & 0.3562 \\
$(1, 2, 4, 8, 16, 24, 48, 96, 168)$ & event50 & 0.7475 & 0.7517 & 0.3769 & 0.3560 \\
\bottomrule
\end{tabular}

  }
  \caption{Absolute test metrics for selected high-performing specifications. Values are reported separately for Fold A and Fold B on the ten percent sample.}
  \label{tab:top_specs_absolute}
\end{table}

\subsection{Compute considerations}
\label{sec:appendix_compute}
The ten percent sample contains about 4.04 million rows.
In the sweep driver logs, the subset of runs that executed in that invocation took 13.9 to 21.2 minutes wall time per run with $n\_jobs=1$.
The same logs recorded a minimum available RAM of 6.45 GiB and minimum free swap of 5.57 GiB during execution.

Target encoding can be computed on the fly from the raw dataset, but this is expensive because it requires repeated hour-shifted aggregations per categorical column.
For the ten percent grid, target encoding features are precomputed and cached in a parquet to amortize this cost.
With target encoding cached, feature generation runtime is dominated by the time aggregation streaming pass and by materializing the dense float32 training matrix for millions of rows.
Peak memory is dominated by large in-memory arrays for the feature matrix and intermediate frames, with additional overhead from per-entity history state.

\subsection{Event-count window sensitivity}
\label{sec:appendix_event_count_sensitivity}
\begin{figure}[t]
  \centering
  \includegraphics[width=\linewidth]{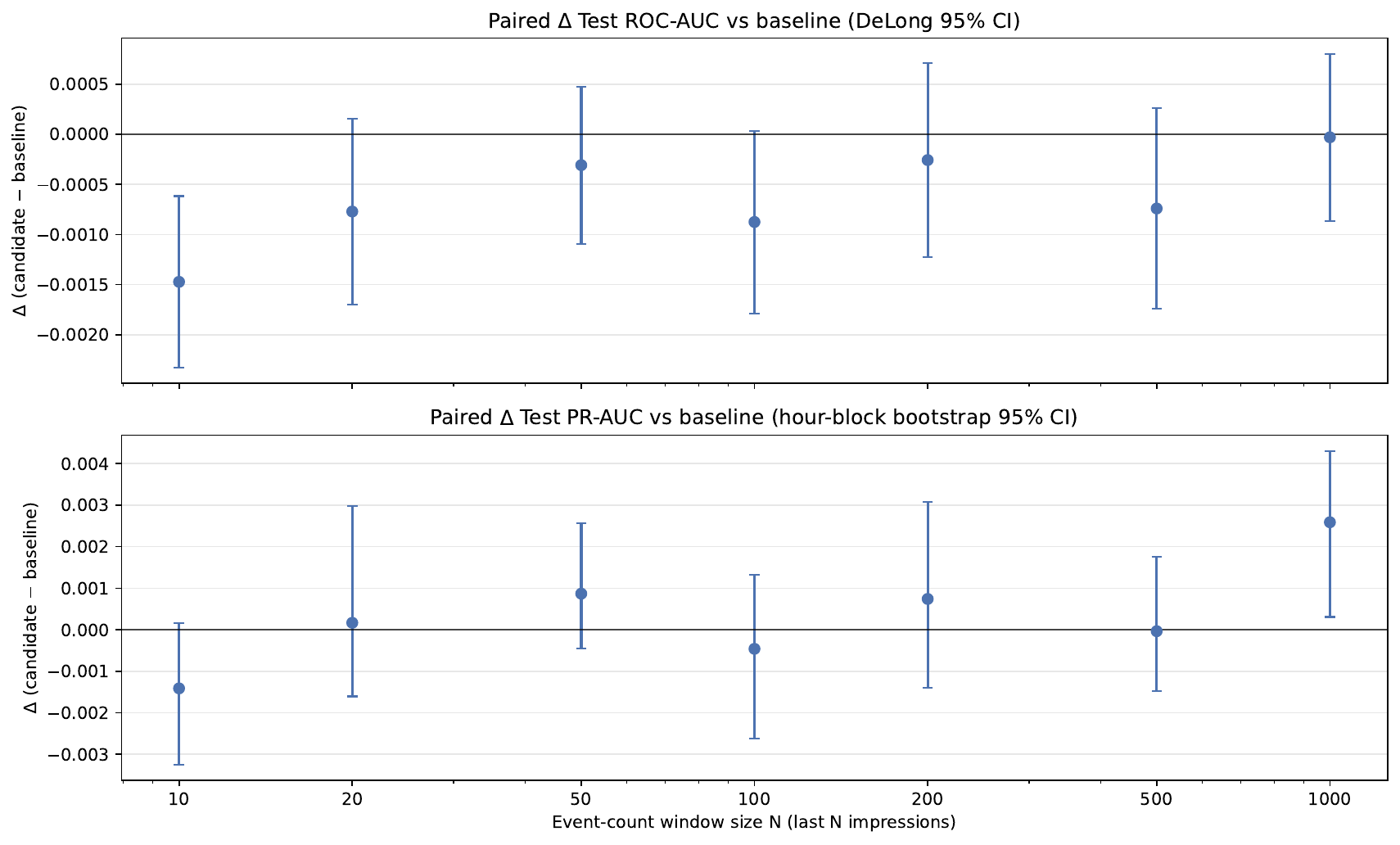}
  \caption{Sensitivity of event-count window size $N$ on a one percent sample with target encoding and length tuple $(1, 6, 24, 48, 168)$. Performance is relatively flat across a wide range of $N$ in this single-split experiment.}
  \label{fig:event_count_sweep}
\end{figure}

Figure \ref{fig:event_count_sweep} reports a small sensitivity sweep that varies the event-count window size $N$ on a one percent sample with target encoding enabled and length tuple $(1, 6, 24, 48, 168)$.
This sweep uses a single split rather than the rolling-tail protocol.

\subsection*{EDA context}
\begin{figure}[t]
  \centering
  \includegraphics[width=\linewidth]{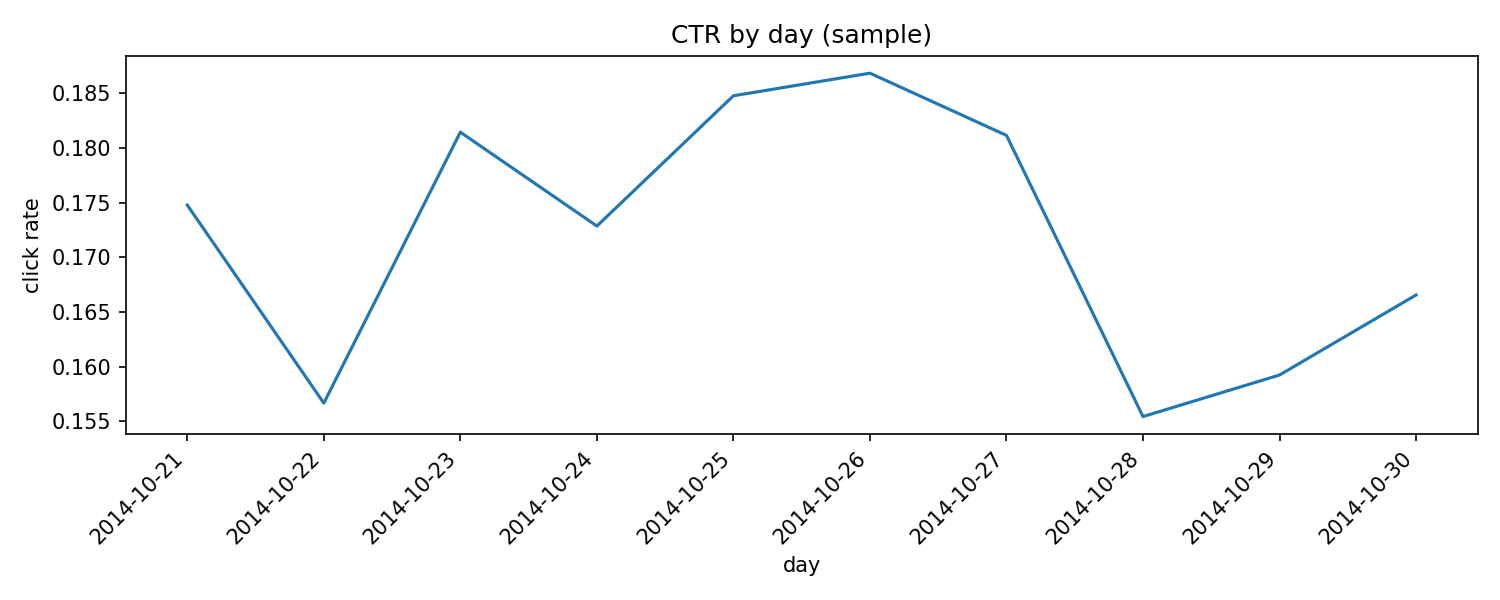}
  \caption{CTR by day in the dataset sample.}
  \label{fig:eda_ctr}
\end{figure}

\begin{figure}[t]
  \centering
  \includegraphics[width=\linewidth]{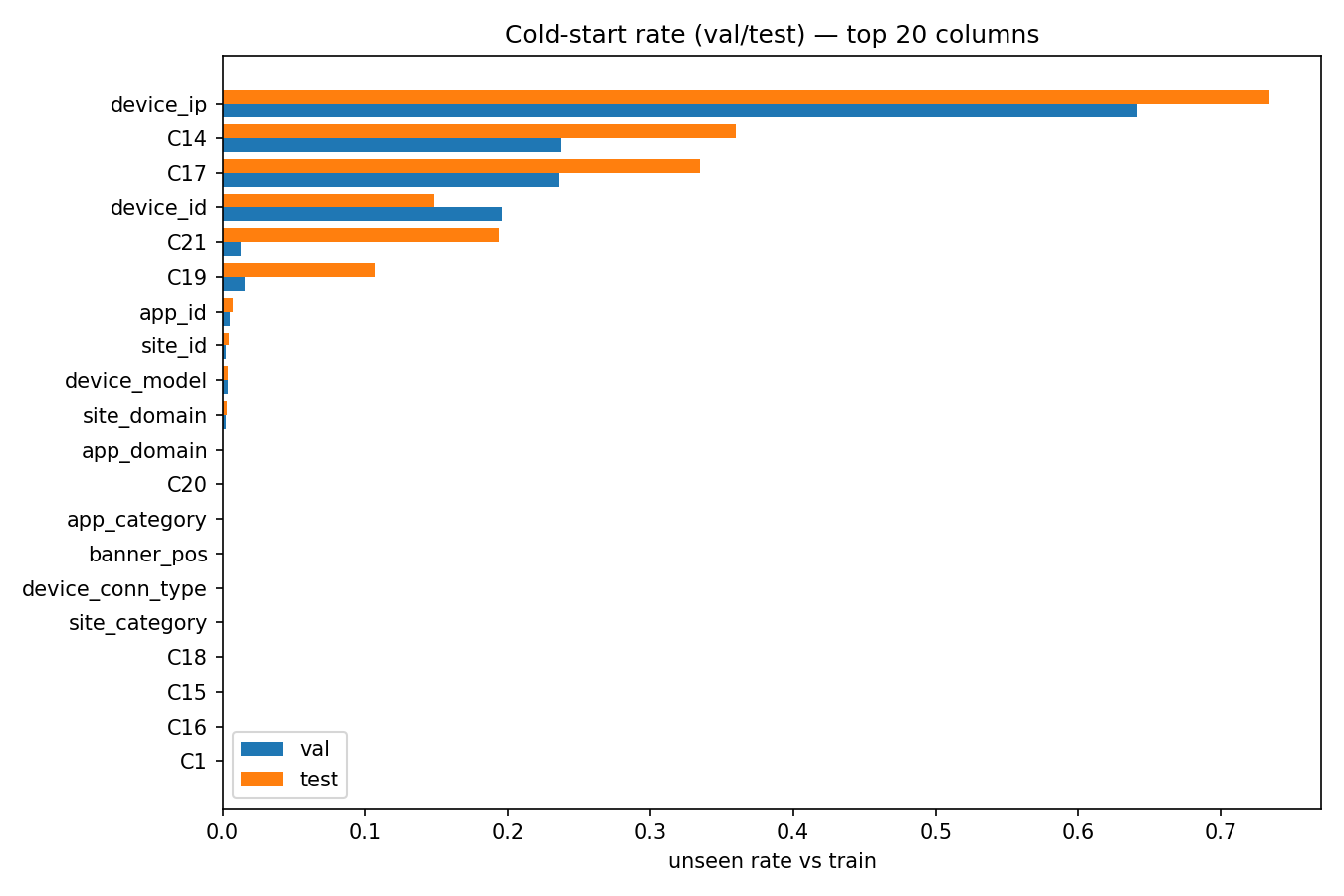}
  \caption{Unseen rate for top categorical features across time.}
  \label{fig:eda_unseen}
\end{figure}

Figure \ref{fig:eda_ctr} shows CTR drift by day.
Figure \ref{fig:eda_unseen} shows an unseen rate summary for top categorical features.
These plots provide context for why out-of-time evaluation and cold-start effects matter.

\subsection*{Full result tables}
This paper includes supplementary tables that report the full design grids and paired inference results.

\begin{figure}[t]
  \centering
  \includegraphics[width=0.92\linewidth]{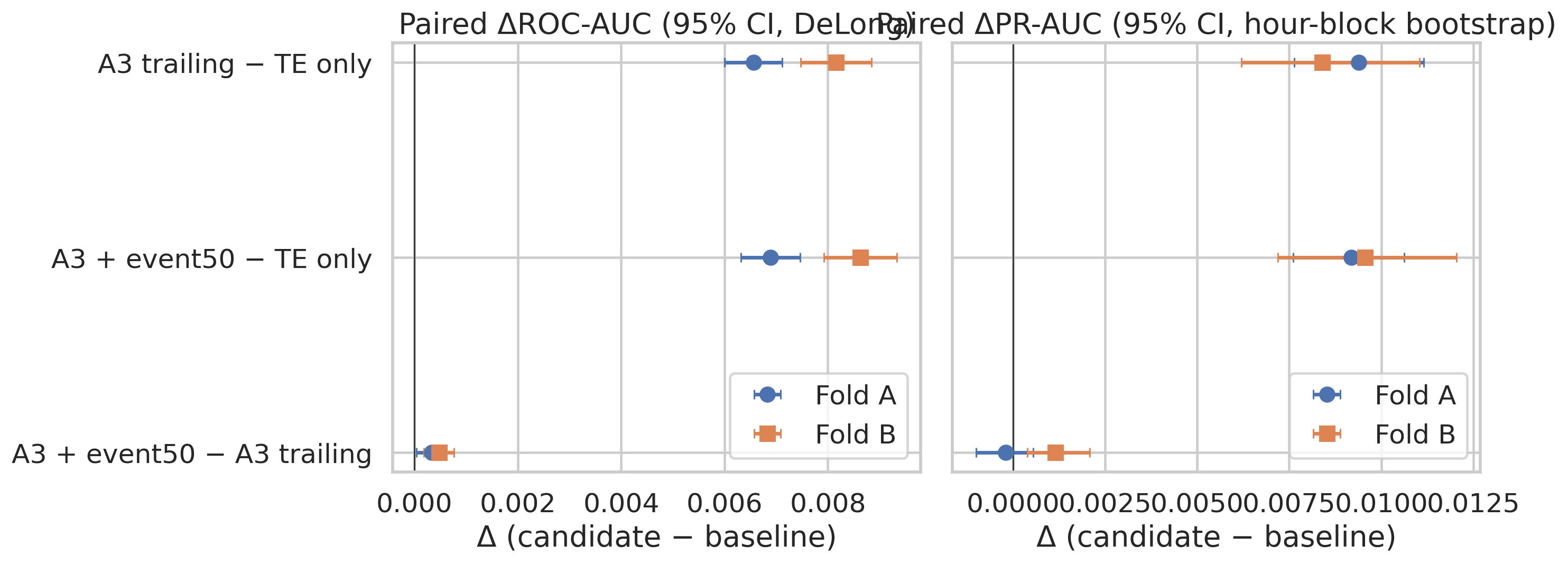}
  \caption{Paired lift from adding time aggregation to a target encoding baseline under no-lookahead evaluation.}
  \label{fig:te_lift_deltas}
\end{figure}

\begin{figure}[t]
  \centering
  \includegraphics[width=\linewidth]{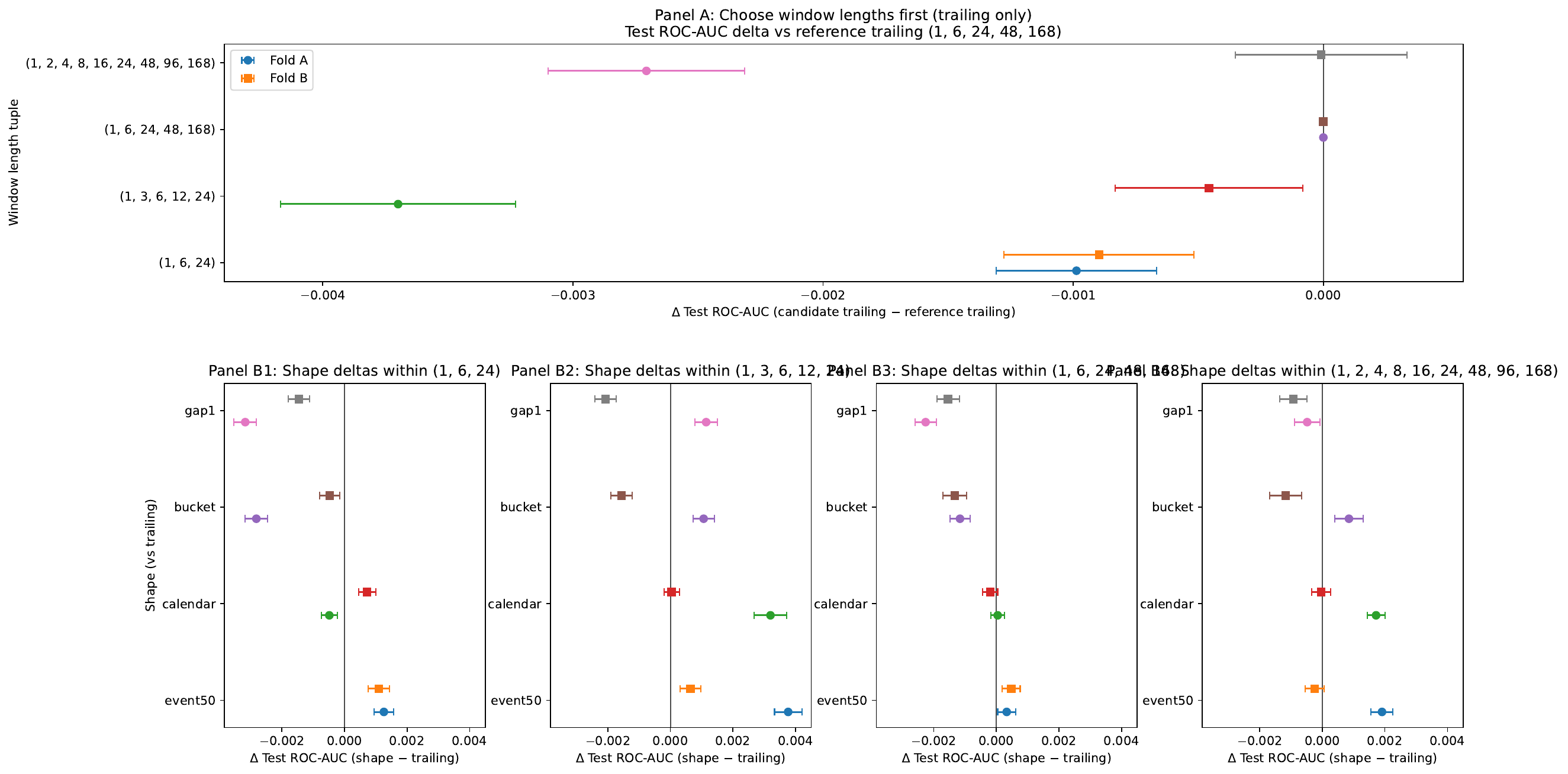}
  \caption{Detailed view of length and shape effects with paired deltas and fold specific estimates.}
  \label{fig:decision_plot}
\end{figure}

\end{document}